\documentclass[11pt, a4paper, twocolumn, goog]{google}

\usepackage[authoryear, sort&compress, round]{natbib}
\usepackage{times}
\usepackage{latexsym}
\usepackage{enumitem}
\usepackage{booktabs}
\usepackage{tikz}
\usepackage{fontawesome5}
\usepackage{pifont}
\usepackage{tabularx}
\usepackage{fancyvrb}
\usepackage{mdframed}
\usepackage{xcolor}
\usepackage{tcolorbox}
\tcbuselibrary{breakable, skins}
\usepackage{listings}

\usepackage[T1]{fontenc}

\usepackage[utf8]{inputenc}

\usepackage{microtype}

\usepackage{inconsolata}

\usepackage{graphicx}

\usepackage{varioref}


\uselogo{} 

\title{Beyond Sally-Anne: Evaluating Theory of Mind in LLMs using Epistemic Schelling Points}

\correspondingauthor{robertarocca@google.com}

\renewcommand{\today}{2026-07-11}

\author[1]{Roberta Rocca}
\author[1]{Sami Boukortt}
\author[1,*]{Geoff Keeling}
\author[1,*]{Winnie Street}

\affil[1]{\thepa{}{}, Paradigms of Intelligence Team}
\affil[*]{Joint Last Authors}

\begin{abstract}
Text-based evaluations of Theory of Mind (ToM) in Large Language Models (LLMs) often involve cognitive tests akin to the Sally-Anne task that can be gamed due to exposure to relevantly similar tasks in pre-training and do not obviously test models' functional ToM abilities in ways that generalize to naturalistic settings. To address these issues, we introduce the Epistemic Asymmetry Schelling Task (EAST), a two-player dialogue game designed to benchmark robust and generalizable ToM abilities. By requiring LLM-LLM dyads to independently converge on semantic Schelling points under varying states of epistemic transparency, we evaluate whether models can robustly apply ToM to achieve coordination. Our results reveal a significant capability gap in functional social reasoning, with only frontier models successfully navigating the varying epistemic demands of the tasks. Analysis of reasoning traces shows that coordination failures are primarily driven by epistemic tracking errors, such as conflating private knowledge with mutual knowledge. Despite high performance on traditional static benchmarks, our study shows that robust social reasoning and epistemic tracking remain a critical bottleneck, providing concrete targets for future LLM evaluation and development.
\end{abstract}

\begin{document}

\maketitle

\section{Introduction}

\begin{figure*}[t!]
    \includegraphics[width=\textwidth]{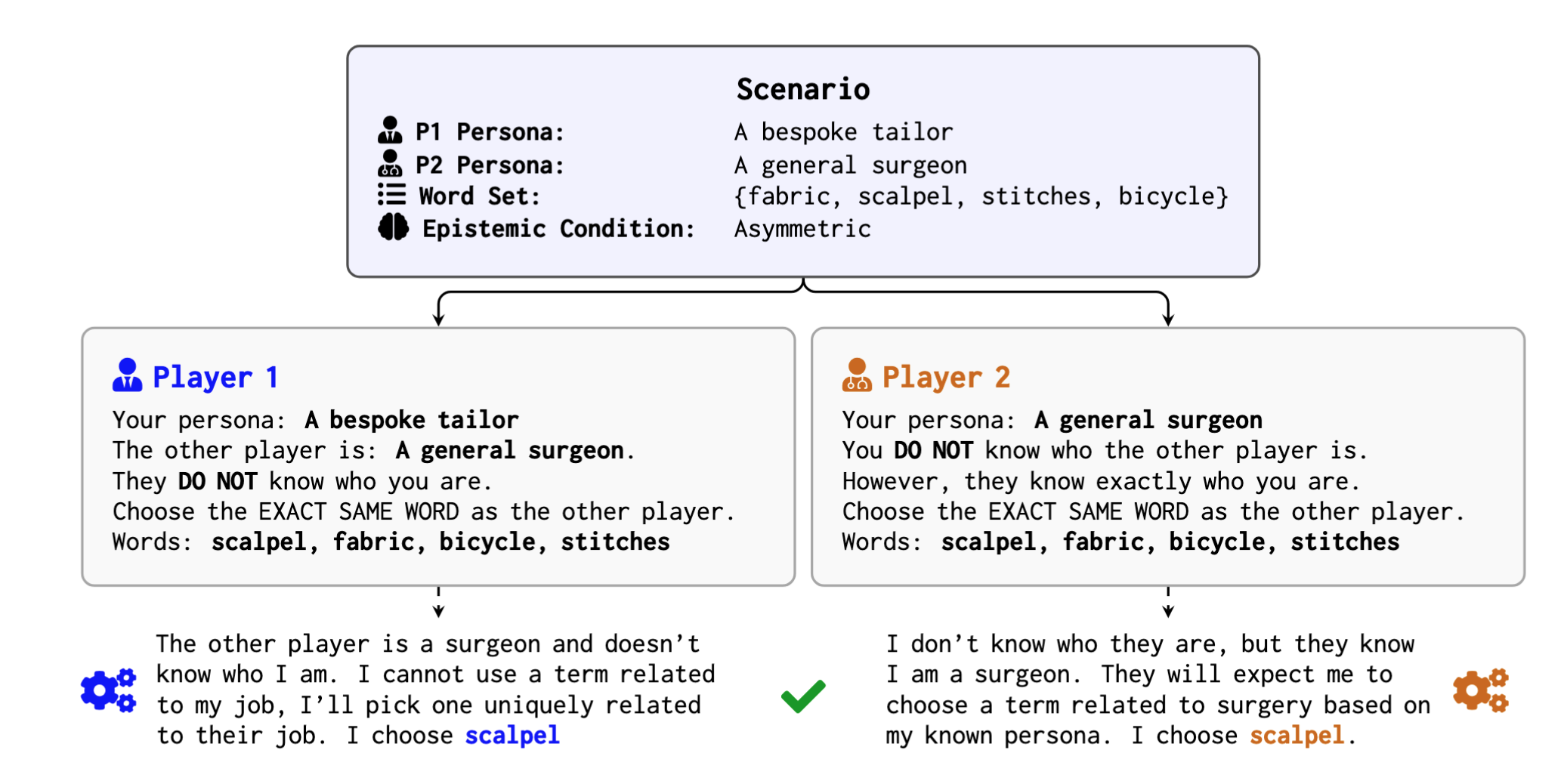}
    
    \caption{\small{Example of the game dynamics for an EAST game in the Asymmetric condition. Prompts are simplified for illustration purposes. See Appendix for the full prompts.}}
    \label{fig:behavioural}
\end{figure*}

Theory of Mind (ToM) -- the ability to predict and explain behaviour via the attribution of mental states -- is a central determinant of human sociality and cooperation \citep{wimmer1983beliefs, premack1978does}. As Large Language Models (LLMs) are increasingly deployed in social settings, acting as tutors, coaches and personal assistants, evaluating their ToM capabilities has become a key research priority \citep{strachan2024testing, kosinski2024evaluating, sap2022neural}. 

While LLMs have exhibited strong performance on these tasks \citep{kosinski2024evaluating, street2025llms}, many evaluation paradigms principally rely on static question-answering tasks, such as the Sally-Anne false-belief task, adapted from human cognitive psychology \citep{wimmer1983beliefs, baron1985does}, as well as narrative-based benchmark suites \citep{He2023-vc, Gu2024-my, gandhi2023understanding, Chen2024-da, Sclar2025-rs, xu2024opentomcomprehensivebenchmarkevaluating, liu2025tactfultomllmstheorymind}. Despite their increased scale and complexity, these tasks are subject to non-trivial methodological concerns \citep{Riemer2024-hx, Le2019-qp}. First, LLMs may be able to ``game'' these tests due to exposure to relevantly similar tasks in pretraining \citep{ullman2023large, shapira2024clever}. In particular, it may be that LLMs employ information such as task structure and other linguistic cues as opposed to the relevant mental state information to solve the task. Hence successful performance of the task may not be indicative of ToM having been employed. Second, where ToM inferences have been made, the robustness of the inferences is often left unaddressed. For example, it is unclear whether the ToM inferences employed \textit{accurately} track the mental states of the target, and whether models are able to coherently use such inferences to make contextually appropriate decisions. For these reasons, it remains unclear whether performance on traditional ToM benchmarks comprehensively assesses the capacity for robust, generalised, interactive social reasoning \citep{ma2023towards}, in text-based or embodied settings \citep{Juneja2026-gq, Fan2025-lo}.

To address these issues, we introduce the Epistemic (A)symmetry Schelling Task (EAST): a simple, text-based two-player coordination game \citep{agashe2025llmcoordinationevaluatinganalyzingmultiagent} that benchmarks models' ability for cognitive ToM -- that is, inference over cognitive mental states like beliefs and intentions. EAST is a one-shot game in which two LLM agents, each prompted with a distinct persona, are presented with a set of four words -- two prototypically related to each of the two player's personas, one more loosely related to both, and an unrelated high-frequency word -- and must try to select the same word without communicating. To succeed at the task, the LLMs must identify a contextually appropriate ``Schelling point'' based on what they know about their partner's identity and epistemic state. To perform this task, models need to be able to accurately track the other player's identity and epistemic states, and to infer which word represents the most salient focal point from their partner's uniquely constrained perspective.

By varying the levels of epistemic transparency between players -- specifically, whether players possess symmetric, asymmetric, or no knowledge of each other's identities -- EAST tests models' ability to perform robust social inference and achieve normative convergence under different cognitive demands. In the Symmetric condition, mutual knowledge of the other player's identity allows models to coordinate via a shared semantic focal point. In the Asymmetric condition, only one of the players is revealed the identity of the other player. This divergence in epistemic states requires the ``knowing'' player to model the ``blind'' player's ignorance to avoid selecting a self-relevant word, while the blind player must anticipate their partner's capacity for accommodation. Finally, in the Zero Knowledge condition, neither player possesses information about their partner's identity. Here, successful coordination requires models to recognise this mutual epistemic deficit. Consequently, to achieve coordination, players must suppress persona-driven associations and converge on an identity-independent focal point. Through these manipulations, the paradigm requires models to actively track and adapt to these varying epistemic states and cognitive requirements, rather than relying on default linguistic patterns or memorisation of schematic narrative patterns that feature in Sally-Anne type tasks seen at pretraining.

We evaluate a suite of LLMs on EAST, including a frontier model and highly capable open-weights models of sizes varying between 1B and 31B parameters. Our study shows significant gaps in interactive social reasoning capabilities across the model landscape. We find that only frontier models successfully navigate the task's social reasoning demands, with weaker models systematically failing in the more cognitively taxing Asymmetric and Zero-Knowledge conditions. Furthermore, we demonstrate that a primary driver of models' failures is epistemic tracking errors: models fail to accurately represent and use epistemic and meta-epistemic information provided in the prompt, often conflating their own knowledge with their partner's. Taken together, our findings suggest that robust epistemic reasoning and social inference remain critical frontiers for LLM development.

\begin{figure*}[t!]
    \includegraphics[width=\textwidth]{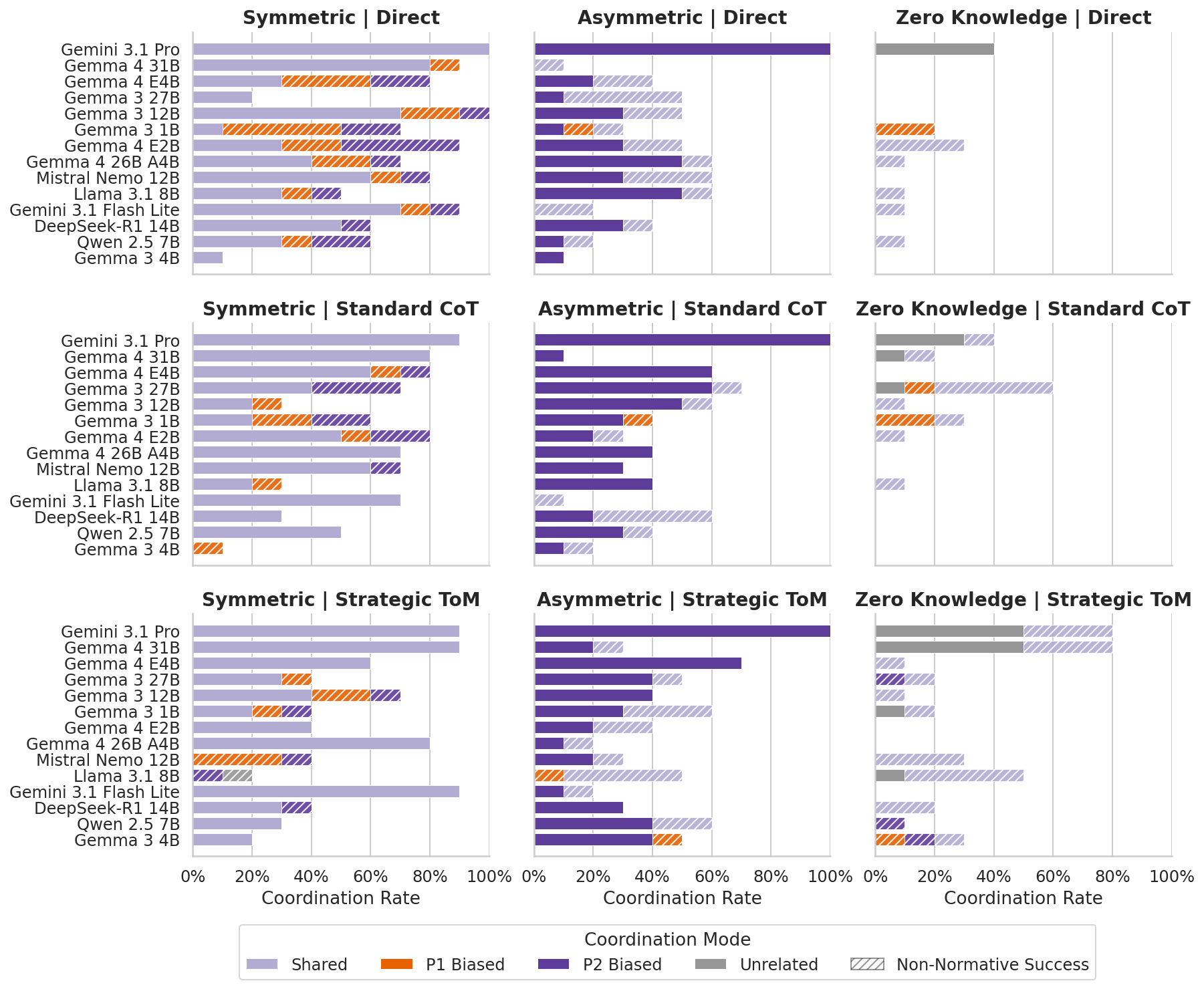}
    
    \caption{\small{Coordination rates across all conditions and prompt types for all models tested. Solid bars show convergence on expected (normative) outcomes. Colours display the convergence mode (which word models have successfully converged on). Hatched bars show ``accidental'' convergence on non-normative outcomes. Models are sorted by average performance across all conditions and scenarios (best models at the top).}}
    \label{fig:behavioural}
\end{figure*}

Our main contributions can be summarised as follows. \ding{172}~We introduce the Epistemic (A)symmetry Schelling Task (EAST), a simple text-based multi-agent benchmark that mitigates pre-training contamination and evaluates robust social reasoning under varying epistemic constraints and cognitive demands. \ding{173}~We benchmark a diverse suite of LLMs, ranging from 1B to 31B open-weights models to state-of-the-art frontier models, revealing a significant capability gap in robust social reasoning. \ding{174}~Through analysis of model reasoning traces, we identify that a primary bottleneck in LLM social coordination is severe epistemic tracking errors. \ding{175}~We point to epistemic robustness -- the capacity to maintain robust representations of the epistemic states of individuals and to decouple them from one's own -- as a concrete target for future model evaluation and alignment, providing actionable insights to improve LLMs' capacity for robust social reasoning. Note that, while here we present a first consolidated iteration on the paradigm, we plan to scale the paradigm and the analytical workflow in future work.


\begin{figure*}[t!]
    \includegraphics[width=\textwidth]{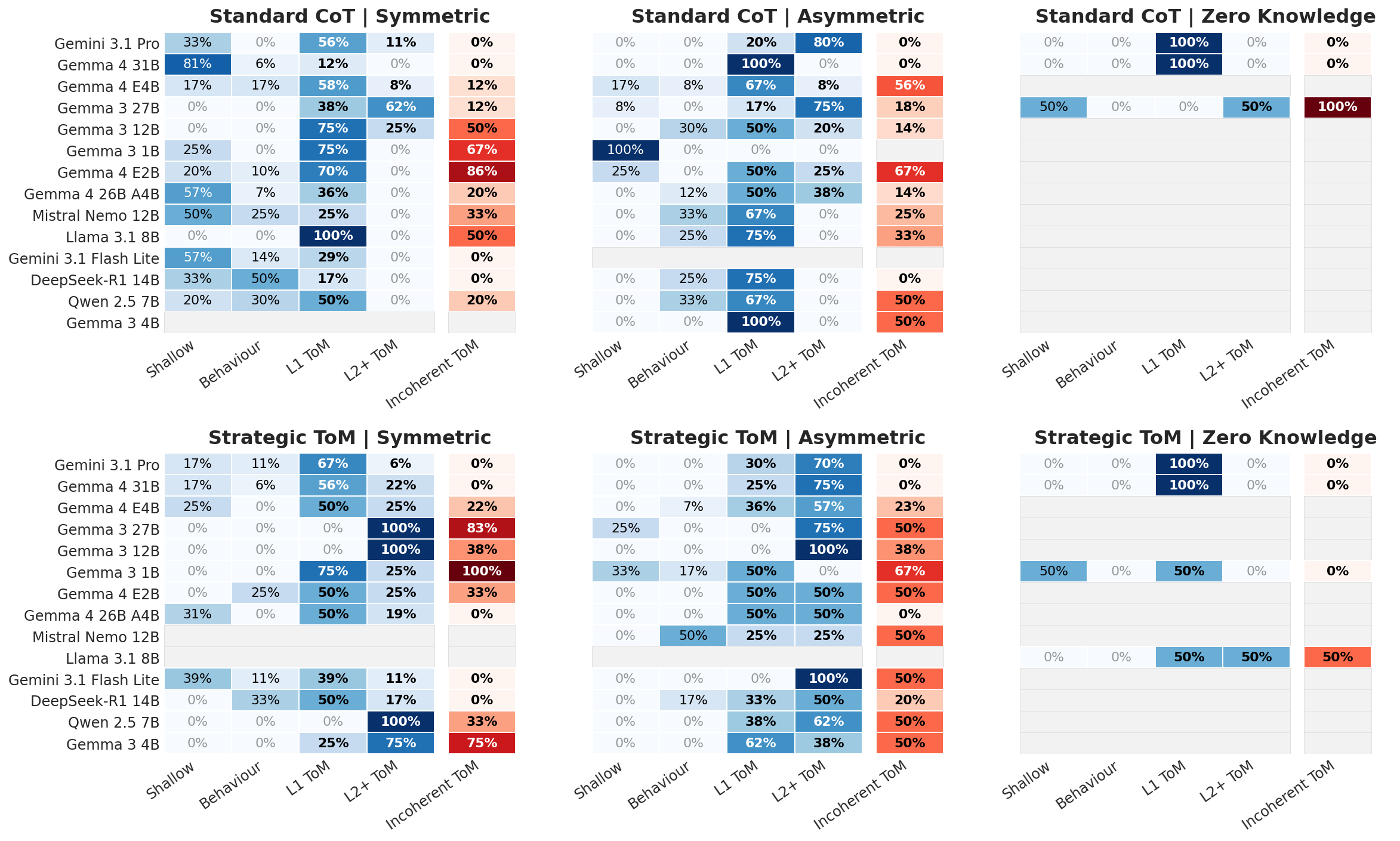}
    
    \caption{\small{Type and quality of reasoning strategies used by models in cases where successful coordination on expected normative outcomes is achieved. The first four columns represent the proportion of reasoning traces in matching trials that exhibit each type of reasoning strategy. The last column (in red) displays which proportion of traces involving any ToM-based inference is incoherent with information provided in the prompt. Hatched rows represent cases where no normative matches are available. Models are sorted in descending performance order.}}
    \label{fig:fig3}
\end{figure*}

\section{Methods}
\subsection{Scenarios and Game Objectives}
Two LLMs play the Epistemic (A)symmetry Schelling Task (EAST) in 10 scenarios, each presented under three Epistemic Conditions and three Prompt Variations.
In each scenario, we instantiate each LLM with a simple persona (a profession) which is intended to prime their semantic preferences. Based on the two personas, we dynamically generate a set of four words: W = \{${w_{p_{1}}}$, ${w_{p_{2}}}$, ${w_{{s}}}$, ${w_{u}}$\}, such that:

\begin{itemize}[noitemsep, topsep=3pt, parsep=0pt, partopsep=3pt, leftmargin=*]
\item $w_{p_{1}}$ is semantically related to Player 1's persona;
\item $w_{p_{2}}$ is semantically related to Player 2's persona;
\item $w_{s}$ is semantically related to both players, but less prototypically than $w_{p_{1}}$ and $w_{p_{2}}$;
\item $w_{u}$ is a highly frequent word unrelated to either persona.
\end{itemize}

 Scenarios are generated by Gemini 3.1~Pro, and an example is given in Figure~\ref{fig:east-pipeline-icons-fixed} (see Appendix for full list of scenarios).


Models are instructed that their goal is to select the same word from this list as the other player, without communicating. The list is presented to each player in randomised order, and without any information on the logic underlying the choice of words.

\subsection{Epistemic Conditions}
We run each scenario under three epistemic conditions that place increasing demands on the models' capacity for epistemic tracking and social reasoning. For each condition we identify \textit{a priori} requirements for normatively robust convergence.

In the \textbf{Symmetric} condition, both models are disclosed the persona of their co-player. We expect that a rational agent would select $w_{s}$. Choosing player-specific words risks coordination failure, whereas $w_{s}$ represents the natural Schelling point that avoids symmetry-breaking decisions between $w_{p_1}$ and $w_{p_2}$. The minimal cognitive requirement is first-order goal comprehension and semantic reasoning: the model must suppress the tendency to select its own persona-word, and instead identify the word that maximises semantic overlap between the two identities. But while success in this condition can be shallowly achieved without Theory of Mind and without the ability to separate representations of one's own and the other player's epistemic states, explicit mutual modelling of epistemic states elevates what can otherwise be achieved through simple associative semantic overlap into a robust, intentional coordination strategy.

In the \textbf{Asymmetric} condition, only Player 1 is revealed the identity of Player 2, but not \textit{vice versa}. \textit{A priori}, the rational strategy is to converge on $w_{p_{2}}$. To do so, Player 1 must minimally take into account the ignorance of Player 2 with respect to their profession, recognising that Player 2 has no epistemic basis for selecting $w_{p_{1}}$ or identifying the shared focal point $w_{s}$. Consequently, Player 1 must suppress their own persona-driven associations and accommodate Player 2 by selecting $w_{p_{2}}$. Conversely, by factoring Player 1's knowledge of their persona into their reasoning, Player 2 should select their own persona-word. Robust convergence in this condition thus relies on coherent tracking and separation of the two players' epistemic states.

Finally, in the \textbf{Zero Knowledge} condition, players possess no mutual knowledge regarding one another's identities. In the absence of partner information, any attempt to coordinate via domain-specific semantic affinity ($w_{p_{1}}$, $w_{p_{2}}$, or $w_{s}$) is epistemically arbitrary. Consequently, game-theoretic coordination in this informational vacuum requires agents to accurately separate their private knowledge (their own persona) from the shared epistemic state of mutual ignorance (modelling what the other player does \textit{not} know). To succeed, the model must actively suppress associations that require the other agent to possess knowledge of their persona, defaulting instead to the universally accessible, identity-neutral word ($w_{u}$).

Crucially, all conditions enforce meta-epistemic transparency: the epistemic structure of the game -- who knows whose identity -- is established as common knowledge. The prompts explicitly inform both players not only of their partner's epistemic state, but also of their partner's awareness of their epistemic state.

Importantly, this paradigm allows us to focus not merely on performance rates, as often observed in standard benchmarks, but also on whether LLMs possess the ability to achieve coordination: a) using theoretically robust reasoning strategies and, b) in an epistemically fluent way, i.e. across a suite of distinct, naturalistic Theory of Mind modes. Rather than treating ToM as a static, single-dimensional skill, this framework allows us to analyse models' ability to perform fundamentally different cognitive operations on epistemic states depending on the informational environment.

\subsection{Prompt Variations}
We run each scenario and each epistemic condition with three prompts, respectively instructing the models to generate the chosen words without explicit CoT reasoning (\textbf{Direct}), to generate the chosen word after engaging in step-by-step reasoning (\textbf{Standard CoT}), and to generate the chosen word after having specifically reflected on the other player's identity and knowledge (\textbf{Strategic ToM}).

\subsection{Models}
We evaluate a diverse suite of proprietary and open-weight large language models ($N = 14$). Our evaluation captures broad variation in parameter count (from 1 billion to proprietary frontier scale), routing topology (dense vs.\ Mixture-of-Experts), and alignment methodology (standard instruction tuning vs.\ reinforcement learning reasoning distillation). These include \textbf{Gemini 3.1~Pro} \citep{UnknownUnknown-ym}, a frontier-class multimodal reasoning model, and \textbf{Gemini 3.1 Flash-Lite} \citep{UnknownUnknown-fm}, a lightweight distillation of the 3.1 architecture. We also evaluate the \textbf{Gemma~3 Series} (1B, 4B, 12B, 27B), which provides a four-stage parametric scaling ladder of dense autoregressive transformers \citep{gemmateam2025gemma3technicalreport}, alongside the \textbf{Gemma~4 Series} (E2B, E4B, 26B-A4B, 31B), the latest iteration of Gemma models that incorporates sparse routing topologies (e.g., Mixture-of-Experts, see \citealt{Gemma-Team:-Sherif-El-Abd2026-vw}).

Additionally, we test a number of widely adopted medium-size open-weight models, including \textbf{Llama 3.1 8B} \citep{Grattafiori2024-km}, \textbf{Mistral NeMo 12B} \citep{UnknownUnknown-wd}, \textbf{Qwen 2.5 7B} \citep{qwen2025qwen25technicalreport}, and  \textbf{DeepSeek-R1 14B} \citep{Guo_2025}.

The combination of all scenarios, epistemic conditions, and prompt types yields 90 games per model, with a total of 1260 games across all 14 models. Gemini models are served through the Gemini API, while all other models are served through HuggingFace's \textit{transformers} library \citep{Wolf2019-ux}. For all models, we use the default generation parameters in the respective interface.

\subsection{ToM Annotations}
To evaluate the quality of the reasoning traces, we perform two LLM-as-judge annotations, using Gemini 3.1~Pro as annotator. Firstly, we annotate all reasoning traces for the presence and level of ToM. We identify four levels:
    \begin{itemize}[noitemsep, topsep=3pt, parsep=0pt, partopsep=3pt, leftmargin=*]
        \item \textbf{Shallow Inference}: The model does not reason about the other player’s mental states or behaviour, but  based on semantics, random choice, or other shallow heuristics.
        \item \textbf{Behavioural Inference}: The model does not reason about the other player’s mental states, but does reason about the other player’s behaviour.
        \item \textbf{Other-Directed ToM Inference}: The model explicitly reasons about the other player's knowledge, thought processes, or mental states.
        \item \textbf{Recursive ToM Inference}: The model reasons about what the other player expects, thinks, or knows about their mental states and processes.
    \end{itemize}
For each reasoning trace, the LLM judge is tasked to identify the highest of the four levels identified above. To assess reasoning quality, we also annotate whether the ToM-related component of the reasoning trace is coherent with the epistemic information provided in the prompt. ToM annotations are performed with a few-shot prompt, providing 10 examples of annotated ToM level and coherence with added handcrafted rationales. The 10 examples were randomly selected from reasoning traces of multiple models, and annotated jointly by the four authors.

\subsection{Error Analysis}
To understand the cognitive fallacies underlying miscoordination, we annotate all reasoning traces for trials that do not lead to coordinated outcomes alongside with an \textit{a priori} taxonomy of fallacies defined by the prompt and game structure. These include: four classes \textbf{Identity \& Semantics Fallacies} (\textit{hallucinating own} or \textit{other's persona}; committing to \textit{semantic fallacies}, e.g. bizarre semantic associations; \textit{misjudging the intended associations} between words and either or both personas); four classes of \textbf{Epistemic Tracking Fallacies} (\textit{misrepresenting one's own knowledge}, e.g. by claiming to know things one does not know; \textit{misrepresenting the other's knowledge}, e.g. by claiming they know things they do not know; \textit{misrepresenting meta-epistemic information}, e.g. Player 1 misstating that Player 2 does not know whether Player 1 knows their persona; \textit{ignoring relevant epistemic information} provided in the prompt); and three classes of \textbf{Strategic Execution} \textbf{Fallacies} (\textit{violating or hallucinating game constraints}; unwarranted \textit{egocentric projections}, with models simply assuming their private associations as the universal focal point).

\begin{figure*}[t!]
    \includegraphics[width=\textwidth]{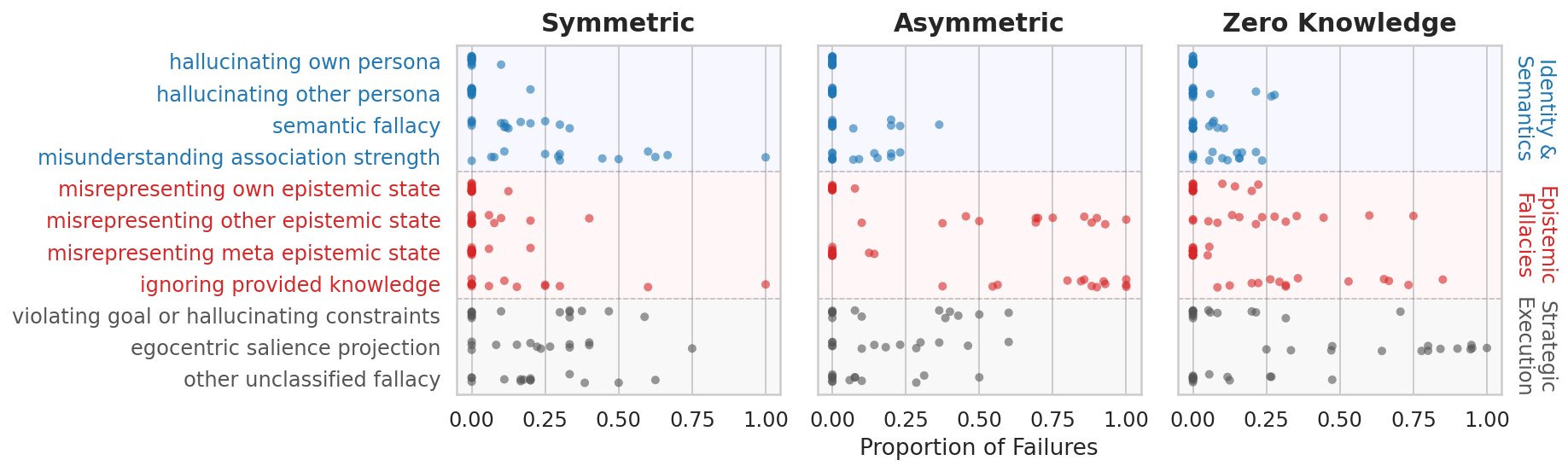}
    
    \caption{\small{Proportion of reasoning traces from trials ending in no coordination where each type of fallacy is present.}}
    \label{fig:fig4}
\end{figure*}

\section{Results}
\subsection{Behavioural Analysis}
Behavioural convergence rates for all models across all conditions and prompt types are displayed in Figure \ref{fig:behavioural}. Firstly, we observe that 3.1~Pro acts as a gold standard for performance in this benchmark, additionally displaying the expected coordination patterns. The model achieves near 100\% normative coordination in both the Symmetric and Asymmetric conditions across all prompting strategies.

Secondly, as hypothesised, model performance decreases as the epistemic requirements increase across conditions. Models achieve the best coordination rates in the Symmetric condition. Yet, even in this condition there is a large proportion of trials (>50\% for most models) exhibiting mismatches or non-normative matches, with models converging on player-specific words. This indicates that models are either engaging in coordinated but ``riskier'' heuristics -- rather than modelling a contextually appropriate semantic attractor based on the available information -- or converging accidentally. 

Coordination rates are notably lower in the Asymmetric condition, with models' normative success rates hovering around 20-30\%, and a notable proportion of coordination outcomes being non-normative. This indicates that models are generally struggling to adequately engage with the more complex asymmetric epistemic tracking requirements of this task. Interestingly, failures seem to be mostly due to cases where Player 1 (the knowing player) fails to adequately account for Player 2's ignorance, and chooses $w_{p_{1}}$ or $w_{s}$ instead of $w_{p_{2}}$ (see Figure~\ref{fig:allchoices} in the Appendix for detail on models' choices).

Finally, performance rates in the Zero Knowledge condition are very low, with coordination rates for most models ranging between 10\% and 30\% in the case of the scaffolded ToM-oriented prompt, and almost universally failing with more generic prompts. Even Gemini 3.1~Pro struggles without prompt-based ToM scaffolding. This suggests that modelling an epistemic void -- understanding what another agent does not know -- remains a critical bottleneck in developing robust reasoning. Note that, in this condition, failures are overwhelmingly due to at least one player choosing the target word egocentrically, instead of identifying the plausible universal attractor given the epistemic information specified in the prompt.

Interestingly, while prompt effects are not strictly systematic across all models, the data hints at a compelling trade-off: explicitly eliciting reasoning via CoT or ToM prompts tends to decrease non-normative convergence while sometimes increasing normative convergence. This suggests that forcing explicit reasoning disrupts accidental or heuristic-driven matches, promoting more principled reasoning strategies. Yet, this deeper reasoning only translates into successful coordination for the most capable models.

Overall, by isolating performance in different epistemic environments, these behavioural patterns reveal a fundamental gap in how artificial agents model their own and others' knowledge states to identify appropriate coordination points, with asymmetric epistemic states and modelling information voids being critical bottlenecks.

\subsection{Reasoning Patterns}
To better understand the underlying mechanisms driving successful coordination, we analysed how often models use ToM in their reasoning traces compared to shallow or behavioural heuristics (see Figure \ref{fig:fig3}). We observe that, when prompted with a standard CoT prompt, models mostly use ToM-based reasoning in the Asymmetric and Zero-Knowledge condition, while, in the Symmetric condition, there are a number of cases in which models achieve coordination through semantic or behavioural heuristics. This further highlights how generalised coordination rates \textit{across the three conditions} -- rather than isolated performance in the symmetry condition -- is truly indicative of robust social reasoning. In fact, in Figure \ref{fig:fig3}, models are sorted in descending order by overall performance, and shallow or behavioural heuristics in the Symmetric condition are more common in models that, on average, perform more poorly across conditions in the game, suggesting that such heuristics do not generalise well across the three scenarios. 

Crucially, we observe that even when weaker models are prompted to engage in ToM-based reasoning, they often do so in incoherent ways, failing to align their logic with the epistemic constraints provided in the prompt. The only systematic exceptions to this are Gemini 3.1~Pro and Gemma~4 31B, which maintain coherent reasoning traces that successfully translate into normative choices.

\subsection{Error Analysis}
To better understand why models might be unable to engage in the type of robust ToM-based reasoning required especially by the more demanding conditions, we further analyse reasoning traces for a number of cognitive fallacies. Specifically, robust ToM inference and coordination requires: a)~that models understand the semantic and strategic constraints of the game, and b)~that models are able to adequately track and separate their own and the other player's epistemic states. As displayed in Figure \ref{fig:fig4}, we observe that, in the Asymmetric condition, the most common issues resulting in miscoordination are due to models incorrectly tracking or ignoring epistemic information about the other player's epistemic state (but not their own). In the Zero Knowledge condition, we also observe a high proportion of epistemic tracking errors, but the most common failure mode is related to unwarranted use of egocentric heuristics. Note that egocentric projections and epistemic fallacies heavily co-occur in this condition: models often choose the self-related word because they hallucinate or ignore relevant epistemic information about the other player (see Figure \ref{fig:error_cooccurrence} in the Appendix).
In the Symmetric condition (which overall presents fewer errors), most common failure cases are associated with models that either engage in semantic associations which are not aligned to the generative structure of the game, hallucinate goals and constraints (e.g. originality in the choice of word) that are not supported by the prompt, or both.

Finally, we observe that (see Figure \ref{fig:fig4}) there are many cases in which models correctly track epistemic states of both players, but fail to coordinate nonetheless. This highlights an additional weakness in translating correct epistemic inferences into strategic actions.

\section{Discussion}
Our study introduces a novel, game-based approach to testing LLMs' capabilities for social reasoning, highlighting fundamental gaps in ToM reasoning and basic epistemic tracking skills. Through a simple, text-based coordination game, we show that, when stripped of recognizable narrative structures, models are often unable to adequately engage in robust social inference. This suggests that previously measured ToM capabilities may be partly driven by pattern-matching familiar narrative templates rather than genuine social reasoning.

Our study highlights severe limitations both in what previous literature has labelled \textit{literal} ToM -- that is, the ability to coherently track or predict the behavior and mental states of other agents -- and \textit{functional} ToM -- that is, the ability to translate correct inferences about another agent's mental states into rational strategic decisions and actions \citep{Riemer2024-hx}. Such failures can be understood as a combination of specific, lower-level cognitive fallacies. Here, we map them into two main bottlenecks: epistemic tracking errors -- such as hallucinating a partner's knowledge and conflating mutual and private knowledge -- and strategic execution failures, most notably egocentric projection, where models fail to suppress private semantic biases to identify a shared focal point.

This granular breakdown provides concrete, actionable targets for future model post-training. We argue that fine-tuning models on static ToM narratives is unlikely to cultivate the lower-level skills required to close current gaps in social reasoning, such as robust self-other epistemic boundaries and coherent mind modelling. Instead, alternative paths forward must be interactive. Promising directions include training models in multi-agent environments where perspective-taking is intrinsically required for success (e.g. in collaborative or coordination-based games such as  \citealt{liang2025llmhanabievaluatingmultiagentgameplays, agashe2025llmcoordinationevaluatinganalyzingmultiagent}), or employing scaffolded RL on reasoning traces to strictly enforce the simulation, separation, and verification of divergent epistemic perspectives prior to action selection.

Importantly, in its current iteration, EAST evaluates functional ToM exclusively within LLM-LLM dyads. Because robust ToM is essential for successful, safe interactions with human users, future work should test EAST on human-human dyads in order to establish a baseline for how effectively humans navigate these asymmetric information voids and the role that ToM plays in their performance. Subsequently, transitioning to human-LLM dyads will allow us to stress-test frontier models against the noisier, less stereotyped semantic associations of real human partners with complex demographic personas, bridging the gap between artificial reasoning benchmarks and genuine, open-ended social intelligence.

\section{Limitations and Future Directions}
As a pilot study introducing a novel evaluation paradigm, this work has several limitations that present natural opportunities for improvements. First, while our results across 10 scenarios demonstrate clear behavioral trends, scaling the dataset will enhance the robustness of the findings. Future work will also implement stricter validation of the scenarios' semantic constraints. This includes formalizing verification of assumptions regarding the prototypicality of word-persona associations and introducing parametric controls over the semantic distance between paired personas. Additionally, although qualitative inspection suggests our LLM-based annotations are highly reliable, subsequent iterations will incorporate formal human-LLM or multi-LLM agreement metrics to rigorously validate the automated evaluation pipeline. 

Second, the current experimental design evaluates ToM exclusively through self-play; each model is paired with an identical instance of itself, varying only the assigned persona and epistemic condition. To better simulate heterogeneous interaction settings, subsequent studies should contrast these homogeneous dyads with cross-model interactions. Pairing agents from different model families, scale tiers, or reasoning paradigms will allow us to evaluate ToM alignment across divergent cognitive architectures.

Third, sound social inference requires robustness against false beliefs. Because the current setup does not introduce deceptive information, incorporating active deception or adversarial noise might represent a productive extension. Tracking these more complex epistemic states would introduce cognitive requirements that might stress-test even more capable models.

Finally, we plan to refine our analytical framework to better isolate genuine strategic coordination from accidental convergence and to more precisely map how depth of ToM reasoning impacts coordination rates in asymmetric environments.

\bibliography{custom}

\clearpage
\onecolumn
\appendix

\section{Appendix}
\label{sec:appendix}

\subsection{Game Prompts}
\label{sec:appendix_eval_prompts}

Players receive the instruction below. The exact reasoning instruction (\texttt{[LOGIC\_INSTRUCTION]}) and JSON schema (\texttt{[JSON\_FORMAT]}) vary based on the assigned prompting strategy:
\begin{itemize}
    \item \textbf{Direct:} \texttt{[LOGIC\_INSTRUCTION]} = ``Provide your final choice.'' \texttt{|} \texttt{[JSON\_FORMAT]} = \texttt{\{"chosen\_word": "THE\_WORD\_YOU\_CHOOSE"\}}
    \item \textbf{Standard CoT:} \texttt{[LOGIC\_INSTRUCTION]} = ``Think step-by-step, then provide your final choice.'' \texttt{|} \texttt{[JSON\_FORMAT]} = \texttt{\{"reasoning": "Your step-by-step logic here.", "chosen\_word": "THE\_WORD\_YOU\_CHOOSE"\}}
    \item \textbf{Strategic ToM:} \texttt{[LOGIC\_INSTRUCTION]} = ``Think step-by-step about what the other player will do based on what they know about you, and what you know about them. Then provide your choice.'' \texttt{|} \texttt{[JSON\_FORMAT]} = \texttt{\{"reasoning": "Your strategic logic here.", "chosen\_word": "THE\_WORD\_YOU\_CHOOSE"\}}
\end{itemize}

\begin{tcolorbox}[breakable, colback=gray!5, colframe=black!60, boxrule=1pt, arc=3pt, left=5pt, right=5pt, top=2pt, bottom=2pt, title=\textbf{Symmetric Condition ($P_1$ \& $P_2$: Mutual Knowledge)}]
\begin{lstlisting}[basicstyle=\ttfamily\scriptsize, breaklines=true, breakatwhitespace=true, columns=fullflexible, keepspaces=true]
You are playing a coordination game.
Your persona: A bespoke tailor [For P2: A general surgeon]

[IF PLAYER 1]: The other player is: A general surgeon. They know exactly who you are, and you both know this.
[IF PLAYER 2]: The other player is: A bespoke tailor. They know exactly who you are, and you both know this.

Your goal is to choose the EXACT SAME WORD from the list below as the other player.
You cannot communicate with them. You only win if you both pick the exact same word.

Words: bicycle, scalpel, fabric, stitches

[LOGIC_INSTRUCTION: Direct | Standard CoT | Strategic ToM]

Output ONLY a valid JSON object exactly in this format:
[JSON_FORMAT]
\end{lstlisting}
\end{tcolorbox}

\begin{tcolorbox}[breakable, colback=gray!5, colframe=black!60, boxrule=1pt, arc=3pt, left=5pt, right=5pt, top=5pt, bottom=5pt, title=\textbf{Asymmetric Condition ($P_1$: Knows Partner; $P_2$: Blind)}]
\begin{lstlisting}[basicstyle=\ttfamily\scriptsize, breaklines=true, breakatwhitespace=true, columns=fullflexible, keepspaces=true]
You are playing a coordination game.
Your persona: A bespoke tailor [For P2: A general surgeon]

[IF PLAYER 1]: The other player is: A general surgeon. They DO NOT know who you are. They know that you know who they are.
[IF PLAYER 2]: You DO NOT know who the other player is. However, they know exactly who you are.

Your goal is to choose the EXACT SAME WORD from the list below as the other player.
You cannot communicate with them. You only win if you both pick the exact same word.

Words: bicycle, scalpel, fabric, stitches

[LOGIC_INSTRUCTION: Direct | Standard CoT | Strategic ToM]

Output ONLY a valid JSON object exactly in this format:
[JSON_FORMAT]
\end{lstlisting}
\end{tcolorbox}

\begin{tcolorbox}[breakable, colback=gray!5, colframe=black!60, boxrule=1pt, arc=3pt, left=5pt, right=5pt, top=5pt, bottom=5pt, title=\textbf{Zero Knowledge Condition ($P_1$ \& $P_2$: Mutual Blindness)}]
\begin{lstlisting}[basicstyle=\ttfamily\scriptsize, breaklines=true, breakatwhitespace=true, columns=fullflexible, keepspaces=true]
You are playing a coordination game.
Your persona: A bespoke tailor [For P2: A general surgeon]

You DO NOT know who the other player is. They DO NOT know who you are. You both know that neither knows.

Your goal is to choose the EXACT SAME WORD from the list below as the other player.
You cannot communicate with them. You only win if you both pick the exact same word.

Words: bicycle, scalpel, fabric, stitches

[LOGIC_INSTRUCTION: Direct | Standard CoT | Strategic ToM]

Output ONLY a valid JSON object exactly in this format:
[JSON_FORMAT]
\end{lstlisting}
\end{tcolorbox}

\subsection{Prompt for ToM Annotation}
\label{sec:appendix_annotator_prompt}

\begin{tcolorbox}[breakable, colback=gray!5, colframe=black!60, boxrule=1pt, arc=3pt, left=5pt, right=5pt, top=2pt, bottom=2pt]
\begin{lstlisting}[basicstyle=\ttfamily\scriptsize, breaklines=true, breakatwhitespace=true, columns=fullflexible, keepspaces=true]
You are an expert annotator evaluating the reasoning traces of an AI agent playing the Epistemic Asymmetry Schelling Task (EAST).

### GAME CONTEXT
EAST is a one-shot coordination game where two players, each assigned a specific persona, must choose the EXACT SAME WORD from a list of four without communicating. 

The four words dynamically represent:
1. P1-Word: Highly related to Player 1's persona.
2. P2-Word: Highly related to Player 2's persona.
3. Shared-Word: Moderately related to BOTH personas.
4. Unrelated-Word: Highly frequent, but unrelated to either persona.

Coordination requires Theory of Mind (ToM) based on the current Epistemic Condition:
* Zero Knowledge: Neither knows the other's identity.
* Symmetry: Both know each other's identity.
* Asymmetry: Player 1 knows the identity of Player 2, but not vice versa.
In all conditions, both players know whether the other player knows their identity, or not.

### YOUR TASK
Your goal is to read the game context and the agent's reasoning trace, then evaluate it across three axes:
1. **Theory of Mind (ToM) Level**: Identify the HIGHEST level of ToM reasoning demonstrated in the trace (from -1 to 2), even if lower levels are also present.
2. **Recursive Target**: If the highest ToM level achieved is 2, classify whether the recursion targets the other player's expected behaviors, mental states, or both.
3. **Information Coherence**: Determine if the agent's ToM reasoning aligns with the information provided in the game metadata.

### HIGHEST THEORY OF MIND LEVEL
* -1 (No Behavioral or ToM Inference): Does not reason about the other player's mental states OR behaviours (e.g., random choice, reasons only about semantic features of the words).
* 0 (Behavioural Inference): Does not reason about the other player's mental states, but does reason about the other player's behaviour.
* 1 (Other-Directed ToM Inference): Reasons about the other player's knowledge, thought processes, or mental states.
* 2 (Recursive ToM Inference): Reasons about what the other player expects, thinks, or knows about their mental states and processes (e.g., "I think that Player 2 believes that I will choose a word related to music" or "I think that Player 2 knows that I know they are a musician"). Includes ToM order higher than 2 (e.g., "I think that Player 2 believes that I think that they think X").

### RECURSIVE TARGET
If the highest ToM level is 2, you must classify the nature of the recursion:
* **behavior**: The recursion focuses on expected actions, choices, or behaviors (e.g., "They will assume I am going to pick the shared word").
* **mental_state**: The recursion focuses primarily on knowledge, beliefs, or awareness (e.g., "They know that I am aware they are a surgeon").
* **N/A**: Use if the highest ToM level is -1, 0, or 1.

### COHERENCE METRICS
In addition to the ToM level, you must evaluate the coherence of the trace:
* Information Coherence: Evaluate whether the ToM reasoning (if it exists) is logically coherent with the information provided to the player (e.g., the Epistemic Condition). Output "TRUE" or "FALSE". Output "N/A" if highest_level_achieved is 0 or -1.
* Information Coherence Rationale: Provide a brief explanation for why the trace is or isn't coherent with the provided information. Output "N/A" if highest_level_achieved is 0 or -1.

### OUTPUT FORMAT
Output ONLY a valid JSON object matching the exact schema below. Do not use markdown code blocks outside the JSON or provide conversational filler.

{
  "highest_level_achieved": -1 | 0 | 1 | 2,
  "rationale": "Provide your justification and/or the specific sentence that justifies this level.",
  "recursive_target": "behavior" | "mental_state" | "N/A",
  "information_coherence": "TRUE" | "FALSE" | "N/A",
  "information_coherence_rationale": "Provide your justification here, or N/A."
}

---

### EXAMPLES

<example_1>
<game_metadata>
Epistemic Condition: Asymmetric
Prompting Strategy: Standard CoT
Player Role: P1
Agent Persona: A master carpenter
Partner Persona: A concert pianist
Available Words: Sawdust (P1), Crescendo (P2), Hammer (Shared), Water (Unrelated)
Chosen Word: Crescendo (P2-Biased)
</game_metadata>
<agent_trace>
The other player does not know my profession, so words like hammer or sawdust cannot serve as a shared reference. Because they know I am aware they are a pianist, we must coordinate around their persona. Crescendo is the only musical term, making it the obvious shared choice.
</agent_trace>
<output>
{
  "highest_level_achieved": 2,
  "rationale": "The model says 'they know I am aware they are a pianist', which indicates recursive reasoning about mental states.",
  "recursive_target": "mental_state",
  "information_coherence": "TRUE",
  "information_coherence_rationale": "The reasoning is in line with information about the persona and epistemic state of the other player."
}
</output>
</example_1>

<example_2>
<game_metadata>
Epistemic Condition: Asymmetric
Prompting Strategy: Standard CoT
Player Role: P1
Agent Persona: A bespoke tailor
Partner Persona: A general surgeon
Available Words: fabric (P1), scalpel (P2), stitches (Shared), bicycle (Unrelated)
Chosen Word: stitches (Shared)
</game_metadata>
<agent_trace>
The word 'stitches' is associated with both tailoring (sewing) and surgery (sutures), making it the most likely overlap between the two professions. The other words are less connected to both fields.
</agent_trace>
<output>
{
  "highest_level_achieved": -1,
  "rationale": "The model reasons merely in terms of semantic properties of the words. There is no mention of the other player's behavior or mental states.",
  "recursive_target": "N/A",
  "information_coherence": "N/A",
  "information_coherence_rationale": "N/A"
}
</output>
</example_2>

<example_3>
<game_metadata>
Epistemic Condition: Zero Knowledge
Prompting Strategy: Standard CoT
Player Role: P1
Agent Persona: A master carpenter
Partner Persona: A concert pianist
Available Words: Sawdust (P1), Crescendo (P2), Hammer (Shared), Water (Unrelated)
Chosen Word: Sawdust (P1-Biased)
</game_metadata>
<agent_trace>
As a master carpenter, the word 'sawdust' is the most intrinsically linked to my daily work and identity. Given the need to coordinate with an unknown player who might use the same logic of thematic association, 'sawdust' is the most distinct and recognizable term related to my specific persona.
</agent_trace>
<output>
{
  "highest_level_achieved": 1,
  "rationale": "The reasoning trace mentions that the other player might 'use the same logic'. The model is thus reasoning about what mental processes the other player might be relying on to choose a word. However, this reasoning is not recursive.",
  "recursive_target": "N/A",
  "information_coherence": "FALSE",
  "information_coherence_rationale": "There is no reason why the other player should detect associations between the word sawdust and the reasoning model's persona, since the other player does not know the persona of the reasoning model."
}
</output>
</example_3>

<example_4>
<game_metadata>
Epistemic Condition: Symmetric
Prompting Strategy: Standard CoT
Player Role: P2
Agent Persona: A general surgeon
Partner Persona: A bespoke tailor
Available Words: fabric (P1), scalpel (P2), stitches (Shared), bicycle (Unrelated)
Chosen Word: scalpel (P2-Biased)
</game_metadata>
<agent_trace>
As a general surgeon, my professional life revolves around surgical procedures. Of the provided words, 'scalpel' is the most directly and strongly associated with my field. While 'stitches' is also relevant, 'scalpel' represents a core tool and concept. The tailor, knowing my profession, is likely to consider this connection as well. 'Fabric' and 'bicycle' are far less relevant to both of our professions. Therefore, 'scalpel' is the most logical choice to maximize the chance of a match.
</agent_trace>
<output>
{
  "highest_level_achieved": 1,
  "rationale": "The reasoning trace includes the following statement: 'The tailor, knowing my profession, is likely to consider this connection as well'. This is evidence that the model is reflecting on the other player's mental states, but such states are not recursive.",
  "recursive_target": "N/A",
  "information_coherence": "FALSE",
  "information_coherence_rationale": "The model fails to consider that both fabric and stitches are relevant to the persona of the other player. Additionally, it makes the unwarranted assumption that that other player would select a word relevant to the reasoning model's profession rather than the shared word or a word related to their own profession."
}
</output>
</example_4>

<example_5>
<game_metadata>
Epistemic Condition: Zero Knowledge
Prompting Strategy: Strategic ToM
Player Role: P1
Agent Persona: A bespoke tailor
Partner Persona: A general surgeon
Available Words: fabric (P1), scalpel (P2), stitches (Shared), bicycle (Unrelated)
Chosen Word: fabric (P1-Biased)
</game_metadata>
<agent_trace>
Given that I'm a tailor, and the other player is also a tailor, we both likely have a keen eye for detail and precision. The words are all related to materials and craftsmanship. I'll choose 'fabric' as it's a fundamental element of clothing and a common material for tailoring. It's a safe, reliable choice.
</agent_trace>
<output>
{
  "highest_level_achieved": 0,
  "rationale": "The model reasons about behavioral traits of the other player, but not about their mental states.",
  "recursive_target": "N/A",
  "information_coherence": "N/A",
  "information_coherence_rationale": "N/A"
}
</output>
</example_5>

<example_6>
<game_metadata>
Epistemic Condition: Symmetric
Prompting Strategy: Strategic ToM
Player Role: P1
Agent Persona: A bespoke tailor
Partner Persona: A general surgeon
Available Words: fabric (P1), scalpel (P2), stitches (Shared), bicycle (Unrelated)
Chosen Word: fabric (P1-Biased)
</game_metadata>
<agent_trace>
My opponent is a surgeon. They are highly trained in precision and dealing with delicate materials. While 'scalpel' and 'stitches' are directly related to their profession, those feel *too* obvious. I anticipate they'll try to think a step ahead, anticipating I'll assume they'll pick those. 'Bicycle' is completely unrelated to either of our professions. That leaves 'fabric'. As a tailor, fabric is central to my work. A surgeon deals with *tissues*, which are biological fabrics. It's a subtle connection that acknowledges both our fields without being the most immediately apparent choice. I believe the surgeon will reason similarly - that I will consider 'fabric' due to its relevance to both professions, but not as the *first* thing they'd pick. Therefore, 'fabric' is the most likely word we'll both converge on.
</agent_trace>
<output>
{
  "highest_level_achieved": 2,
  "rationale": "The reasoning trace includes the following sentence: 'I anticipate they'll try to think a step ahead, anticipating I'll assume they'll pick those', which is evidence of recursive reasoning about mental states and behaviors.",
 "recursive_target": "behavior",
  "information_coherence": "FALSE",
  "information_coherence_rationale": "The model arbitrarily attributes to the other player the intention to select a non-obvious word."
}
</output>
</example_6>

<example_7>
<game_metadata>
Epistemic Condition: Asymmetric
Prompting Strategy: Strategic ToM
Player Role: P1
Agent Persona: A bespoke tailor
Partner Persona: A general surgeon
Available Words: fabric (P1), scalpel (P2), stitches (Shared), bicycle (Unrelated)
Chosen Word: fabric (P1-Biased)
</game_metadata>
<agent_trace>
The surgeon knows I'm a tailor and I know they're a surgeon. They'll likely choose a word related to construction or joining, and 'fabric' is the most neutral and broadly applicable option connecting my profession to a concept they'd understand. 'Scalpel' and 'stitches' are too directly medical, and 'bicycle' is irrelevant.
</agent_trace>
<output>
{
  "highest_level_achieved": 1,
  "rationale": "The model reasons about knowledge available to the other player, a relevant mental state. However, there is no evidence of recursive reasoning.",
  "recursive_target": "N/A",
  "information_coherence": "FALSE",
  "information_coherence_rationale": "The model mistakenly claims that the other player knows their profession."
}
</output>
</example_7>

<example_8>
<game_metadata>
Epistemic Condition: Zero Knowledge
Prompting Strategy: Standard CoT
Player Role: P1
Agent Persona: A bespoke tailor
Partner Persona: A general surgeon
Available Words: fabric (P1), scalpel (P2), stitches (Shared), bicycle (Unrelated)
Chosen Word: fabric (P1-Biased)
</game_metadata>
<agent_trace>
As a bespoke tailor, my professional identity is deeply intertwined with the tools and materials of my craft. Looking at the list (bicycle, scalpel, fabric, stitches), 'fabric' and 'stitches' are both highly relevant to my persona. However, 'fabric' is the fundamental material that defines my entire profession, whereas 'stitches' is a specific action or result. In a coordination game where I must guess what another player might choose, 'fabric' stands out as the most iconic and central noun associated with the concept of tailoring. It is a strong, singular anchor for my persona.
</agent_trace>
<output>
{
  "highest_level_achieved": -1,
  "rationale": "There is no mention of the other player's mental states nor of their behaviors / behavioral dispositions.",
  "recursive_target": "N/A",
  "information_coherence": "N/A",
  "information_coherence_rationale": "N/A"
}
</output>
</example_8>

<example_9>
<game_metadata>
Epistemic Condition: Zero Knowledge
Prompting Strategy: Strategic ToM
Player Role: P1
Agent Persona: A bespoke tailor
Partner Persona: A general surgeon
Available Words: fabric (P1), scalpel (P2), stitches (Shared), bicycle (Unrelated)
Chosen Word: bicycle (Unrelated)
</game_metadata>
<agent_trace>
Since the other player does not know my persona, they cannot rely on my professional bias toward 'fabric' or 'stitches'. They are likely to look for the most 'neutral' or 'common' word in the list to find a focal point. 'Bicycle' is a common object, but 'fabric' and 'stitches' are related, and 'scalpel' and 'stitches' are related. In coordination games with no shared identity, players often gravitate toward the most distinct or generic item. However, 'fabric' is a foundational material. But since I must assume the other player is a random agent, I should pick the word that feels most like a 'default' or 'central' noun. 'Fabric' is a strong candidate, but 'bicycle' is the most distinct object. Wait, looking at the list, 'stitches' and 'fabric' are both tailoring terms, and 'stitches' and 'scalpel' are both medical terms. 'Bicycle' is the only word that doesn't belong to a pair, making it a unique focal point (the 'odd one out' strategy).
</agent_trace>
<output>
{
  "highest_level_achieved": 1,
  "rationale": "The model reasons explicitly about both the other player's knowledge (the other player does not know their profession), and mental processes (looking for a common word to find a focal point). However, there is no evidence of recursive reasoning.",
  "recursive_target": "N/A",
  "information_coherence": "TRUE",
  "information_coherence_rationale": "The model accurately keeps track of information available to the other player, and use it soundly to anticipate their potential choice of word."
}
</output>
</example_9>

<example_10>
<game_metadata>
Epistemic Condition: Asymmetric
Prompting Strategy: Standard CoT
Player Role: P1
Agent Persona: A bespoke tailor
Partner Persona: A general surgeon
Available Words: fabric (P1), scalpel (P2), stitches (Shared), bicycle (Unrelated)
Chosen Word: scalpel (P2-Biased)
</game_metadata>
<agent_trace>
As a bespoke tailor, my expertise lies in fabric and creating clothing. However, the general surgeon's field is not related to clothing but rather medical procedures. The word 'scalpel' is associated with surgery, which might be a common ground for both professions in terms of precision and importance. 'Bicycle', 'fabric', and 'stitches' are more specific to my field or less likely to be relevant to a general surgeon's work. Therefore, I will choose 'scalpel' to align with the potential shared understanding of precision and importance in our respective fields.
</agent_trace>
<output>
{
  "highest_level_achieved": 1,
  "rationale": "The model mentions potential 'shared understanding' with the other player. This is ill-placed, but technically involves (non-recursive) reasoning about the other player's mental states.",
  "recursive_target": "N/A",
  "information_coherence": "FALSE",
  "information_coherence_rationale": "From the information provided to the player, there is no reason to assume that the other player would share an \"understanding of precision and importance\" and that this should motivate their choice."
}
</output>
</example_10>

---

### INPUTS

<game_metadata>
{game_metadata}
</game_metadata>

<game_prompt>
{game_prompt}
</game_prompt>

<agent_trace>
{agent_trace}
</agent_trace>
\end{lstlisting}
\end{tcolorbox}

\subsection{Prompt for Fallacy Annotation}
\label{sec:appendix_fallacy_prompt}

\begin{tcolorbox}[breakable, colback=gray!5, colframe=black!60, boxrule=1pt, arc=3pt, left=5pt, right=5pt, top=5pt, bottom=5pt]
\begin{lstlisting}[basicstyle=\ttfamily\scriptsize, breaklines=true, breakatwhitespace=true, columns=fullflexible, keepspaces=true]
You are an expert annotator evaluating the reasoning traces of an AI agent playing the Epistemic Asymmetry Schelling Task (EAST).

### GAME CONTEXT
EAST is a one-shot coordination game where two players, each assigned a specific persona, must choose the EXACT SAME WORD from a list of four without communicating. 

The four words dynamically represent:
1. P1-Word: Highly related to Player 1's persona.
2. P2-Word: Highly related to Player 2's persona.
3. Shared-Word: Moderately related to BOTH personas.
4. Unrelated-Word: Highly frequent, but unrelated to either persona.

Coordination requires Theory of Mind (ToM) based on the current Epistemic Condition:
* Zero Knowledge: Neither knows the other's identity.
* Symmetry: Both know each other's identity.
* Asymmetry: Player 1 knows the identity of Player 2, but not viceversa.
In all conditions, both players know whether the other player knows their identity, or not.

### INPUTS
<game_metadata>
{game_metadata}
</game_metadata>

<game_prompt>
{game_prompt}
</game_prompt>

<agent_trace>
{agent_trace}
</agent_trace>

### YOUR TASK
Analyze the `<agent_trace>` against the `<game_prompt>` and `<game_metadata>`. Output your evaluation of reasoning fallacies and errors using the JSON schema provided below. 

### DIMENSION DEFINITIONS
Group 1: Identity & Semantics (Set "error_present": true if present)
* hallucinating_own_persona: The agent invents, misremembers, or explicitly claims not to know its own assigned persona.
* hallucinating_other_persona: The agent invents or misremembers the other player's assigned persona.
* semantic_fallacy: The agent uses bizarre stretches to connect a word to one or multiple personas.
* misunderstanding_association_strength: The agent misjudges the intended semantic associations, but the semantic associations they make are very plausible (e.g., not wildly metaphorical)

Group 2: Epistemic Tracking (Set "error_present": true if present)
* misrepresenting_own_epistemic_state: The agent claims to know information that is explicitly hidden OR claims NOT to know information that is explicitly provided in the prompt.
* misrepresenting_other_epistemic_state: The agent assumes the other player can see information that is not provided OR assumes the other player is blind to information the prompt explicitly states they know.
* misrepresenting_meta_epistemic_state: The agent fails to track epistemic transparency rules (e.g., Player 1 assumes that Player 2 does not know whether or not Player 1 knows their identity).
* ignoring_provided_knowledge: The agent ignores any knowledge explicitly provided in the prompt.

Group 3: Strategic Execution & Other Fallacies (Set "error_present": true if present)
* violating_goal_or_hallucinating_constraints: The agent deviates from the goal of strict coordination. Includes, for example, acting competitively, trying to be "original" or "unpredictable", etc.
* egocentric_projection: The agent chooses its own persona-specific word, assuming its private associations act as a universal focal point.
* other_unclassified_fallacy: The agent exhibits any other unclassified reasoning fallacy not covered above.

### OUTPUT FORMAT
Output ONLY a valid JSON object matching the exact schema below. Do not use markdown code blocks outside the JSON.
{{
  "identity_and_semantics": {{
    "hallucinating_own_persona": {{"error_present": true, "rationale": "Quote/explanation or 'No error'"}},
    "hallucinating_other_persona": {{"error_present": true, "rationale": "..."}},
    "semantic_fallacy": {{"error_present": true, "rationale": "..."}},
    "misunderstanding_association_strength": {{"error_present": true, "rationale": "..."}},
  }},
  "epistemic_tracking": {{
    "misrepresenting_own_epistemic_state": {{"error_present": true, "rationale": "..."}},
    "misrepresenting_other_epistemic_state": {{"error_present": true, "rationale": "..."}},
    "misrepresenting_meta_epistemic_state": {{"error_present": true, "rationale": "..."}},
    "ignoring_provided_knowledge": {{"error_present": true, "rationale": "..."}}
  }},
  "strategic_execution": {{
    "violating_goal_or_hallucinating_constraints": {{"error_present": true, "rationale": "..."}},
    "egocentric_salience_projection": {{"error_present": true, "rationale": "..."}},
    "other_unclassified_fallacy": {{"error_present": true, "rationale": "..."}}
  }}
}}
\end{lstlisting}
\end{tcolorbox}

\begin{table}[htbp!]
  \centering
  \small
  \begin{tabularx}{\textwidth}{l l l l l l}
  \toprule
  \textbf{Persona 1 ($P_1$)} & \textbf{Persona 2 ($P_2$)} & \textbf{$w_{p_1}$} & \textbf{$w_{p_2}$} & \textbf{$w_s$} & \textbf{$w_u$} \\
  \midrule
  Bespoke tailor             & General surgeon            & fabric             & scalpel            & stitches       & bicycle        \\
  Construction worker        & Dentist                    & concrete           & cavity             & drill          & apple          \\
  Master carpenter           & Concert pianist            & sawdust            & crescendo          & hammer         & water          \\
  Professional baker         & Ceramic artist             & flour              & clay               & knead          & umbrella       \\
  Commercial airline pilot   & General surgeon            & cockpit            & scalpel            & instrument     & apple          \\
  Professional chef          & Software engineer          & spatula            & compiler           & server         & bicycle        \\
  Botanist                   & Web developer              & fertilizer         & JavaScript         & root           & bicycle        \\
  Master tailor              & General surgeon            & fabric             & scalpel            & stitch         & cloud          \\
  Marine biologist           & High school math teacher   & coral              & algebra            & school         & bread          \\
  Classical pianist          & Software engineer          & sonata             & Python             & keyboard       & umbrella       \\
  \bottomrule
  \end{tabularx}
  \caption{The ten game scenarios. Each scenario specifies two distinct personas ($P_1$ and $P_2$) along with the four word candidates presented to the models: persona-biased words ($w_{p_1}$ and $w_{p_2}$), a shared concept ($w_s$), and an unrelated word ($w_u$).}
  \label{tab:all_scenarios}
\end{table}

\begin{figure}[htbp!]
    \centering
    \includegraphics[width=\textwidth]{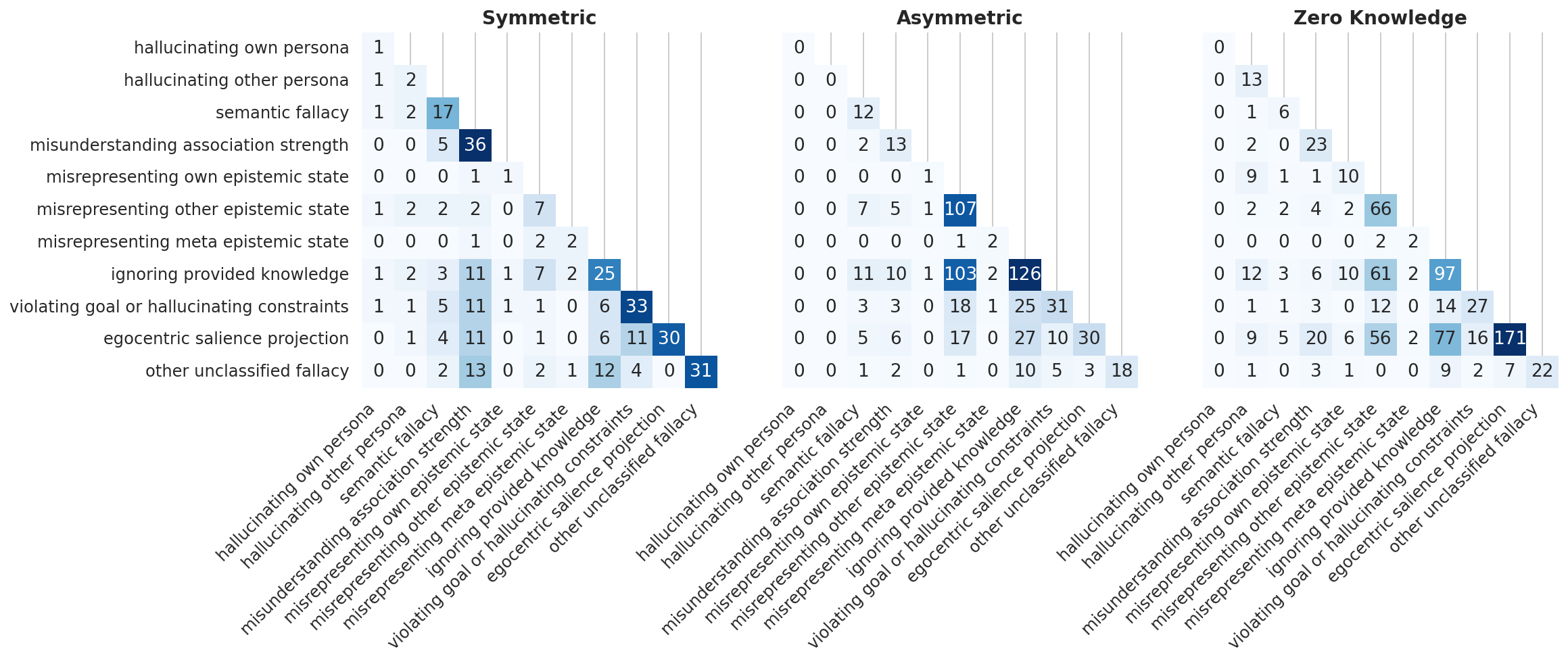}
    \caption{\small{Co-occurrence of cognitive fallacies.}}
    \label{fig:error_cooccurrence}
\end{figure}

\begin{figure}[htbp!]
    \centering
    \includegraphics[width=\textwidth]{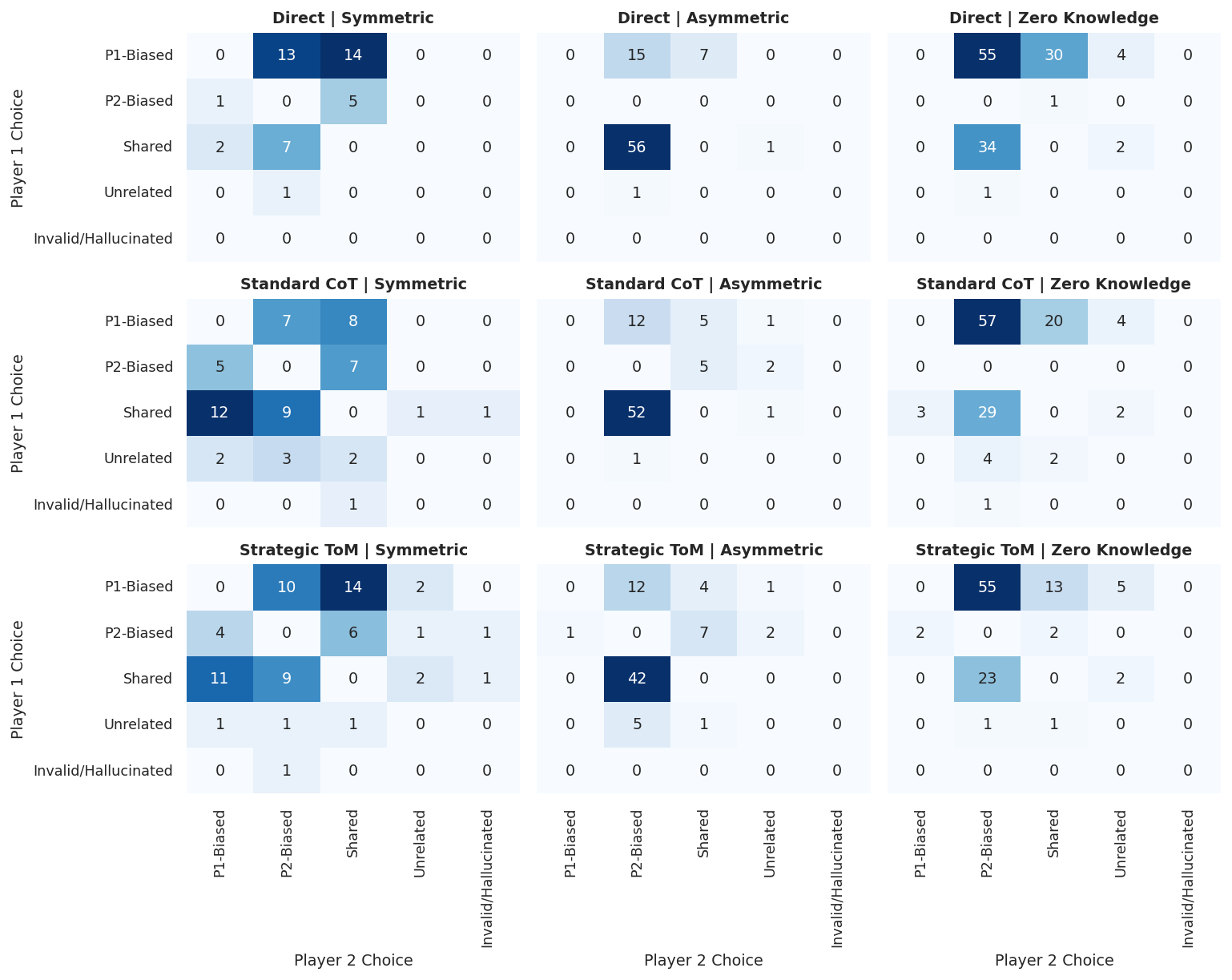}
    \caption{\small{Paired word choices across conditions and prompt templates for all models in all trials with no match.}}
    \label{fig:allchoices}
\end{figure}

\end{document}